\documentclass{article}

\usepackage{arxiv}

\usepackage[utf8]{inputenc} % allow utf-8 input
\usepackage[T1]{fontenc}    % use 8-bit T1 fonts
\usepackage{hyperref}       % hyperlinks
\usepackage{url}            % simple URL typesetting
\usepackage{booktabs}       % professional-quality tables
\usepackage{amsfonts}       % blackboard math symbols
\usepackage{nicefrac}       % compact symbols for 1/2, etc.
\usepackage{microtype}      % microtypography
\usepackage{lipsum}		% Can be removed after putting your text content
\usepackage{graphicx}
\usepackage{natbib}
\usepackage{doi}

\title{ScanBank: A Benchmark Dataset for Figure Extraction from Scanned Electronic Theses and Dissertations}

\date{}

\author{ \href{https://orcid.org/0000-0002-8522-2926}{\includegraphics[scale=0.06]{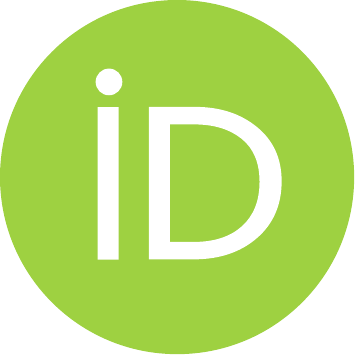}\hspace{1mm}Sampanna Yashwant Kahu}\\
	Department of Electrical and Computer Engineering\\
	Virginia Polytechnic Institute and State University\\
	Blacksburg, VA 24061 \\
	\texttt{sampanna@vt.edu} \\
	\And
	\href{https://orcid.org/0000-0002-8307-8844}{\includegraphics[scale=0.06]{orcid.pdf}\hspace{1mm}William A. Ingram} \\
	Department of Computer Science\\
	Virginia Polytechnic Institute and State University\\
	Blacksburg, VA 24061 \\
	\texttt{waingram@vt.edu} \\
	\And
	\href{https://orcid.org/0000-0003-1447-6870}{\includegraphics[scale=0.06]{orcid.pdf}\hspace{1mm}Edward A. Fox} \\
	Department of Computer Science\\
	Virginia Polytechnic Institute and State University\\
	Blacksburg, VA 24061 \\
	\texttt{fox@vt.edu} \\
	\And
	\href{https://orcid.org/0000-0003-0173-4463}{\includegraphics[scale=0.06]{orcid.pdf}\hspace{1mm}Jian Wu} \\
	Department of Computer Science\\
	Old Dominion University\\
	Norfolk, VA 23529 \\
	\texttt{jwu@cs.odu.edu} \\
}

\renewcommand{\headeright}{}
\renewcommand{\undertitle}{}
\renewcommand{\shorttitle}{}

\hypersetup{
pdftitle={ScanBank: A Benchmark Dataset for Figure Extraction from Scanned Electronic Theses and Dissertations},
pdfsubject={cs.DL, cs.CV, cs.LG},
pdfauthor={Sampanna Yashwant Kahu, William A. Ingram, Edward A. Fox, Jian Wu},
pdfkeywords={Dataset, Digital Libraries, Deep Neural Networks, YOLOv5, Figure Extraction, Electronic Theses and Dissertations},
}

\begin{document}
\maketitle

\begin{abstract}
% What is the problem? 
We focus on electronic theses and dissertations (ETDs), aiming to improve access and expand their utility, since more than 6 million are publicly available, and they constitute an important corpus to aid research and education across disciplines.
The corpus is growing as new born-digital documents are included, and since
millions of older theses and dissertations have been converted to digital form to be disseminated electronically in institutional repositories.
In ETDs, as with other scholarly works,
% Figure extraction is an important aspect of scholarly text and data mining because 
figures and tables can communicate a large amount of information in a concise way.
Although methods have been proposed for extracting figures and tables from born-digital PDFs, they do not work well with scanned ETDs.  
% Why is the problem a problem? 
Considering this problem, our assessment of state-of-the-art figure extraction systems is that the reason they do not function well on scanned PDFs is that they  have only been trained on born-digital documents.
% What is your secret sauce?
To address this limitation, we present ScanBank, a new dataset containing 10 thousand scanned page images, manually labeled by humans as to the presence of the 3.3 thousand figures or tables found therein.
We use this dataset to train a deep neural network model based on YOLOv5 to accurately extract figures and tables from scanned ETDs. 
We pose and answer important research questions aimed at finding better methods for figure extraction from scanned documents.
One of those concerns the value for training, of data augmentation techniques applied to born-digital documents which are used to train models better suited for figure extraction from scanned documents.
To the best of our knowledge, ScanBank is the first manually annotated dataset for figure and table extraction for scanned ETDs. 
A YOLOv5-based model, trained on ScanBank, outperforms existing comparable open-source and freely available baseline methods by a considerable margin.
\end{abstract}

% keywords can be removed
% Dataset, Digital Libraries, Deep Neural Networks, YOLOv5, Figure Extraction, Electronic Theses and Dissertations

% --------------------------------------------------------------------
% --------------------------------------------------------------------
% --------------------------- Introduction ---------------------------
% --------------------------------------------------------------------
% --------------------------------------------------------------------

\section{Introduction}
Over the past decade, deep learning techniques have significantly boosted the accuracy of object detection and classification in natural images \cite{koudas2020video,laroca2018robust}. 
Document objects like figures and tables contain important information.
Their automatic identification and extraction from PDF files is key to enhancing computational access to scholarly works.
This facilitates important operations such as semantic parsing, searching, and summarizing.
% The ability to extract figures and tables from scientific documents can facilitate key use cases such as semantic parsing, summarization, and indexing.
Our research focuses on Electronic Theses and Dissertations (ETDs), aiming to improve access and expand their utility.
Since more than 6 million ETDs are publicly available, they constitute an important corpus to aid research and education across disciplines.
Beginning with Virginia Tech in 1997, graduate programs all over the world allow (or often mandate) electronic submission of an ETD as a requirement for graduation.
University libraries often provide public access to digital libraries of ETDs.
Some universities scan older theses and dissertations to provide electronic access to these older works.
For instance, Virginia Tech's ETD collection dates back to the year 1903.
Most ETDs before the late 1990s are scanned versions of physical copies.
Our work aims at identifying and extracting figures and tables  from scanned ETDs.
For brevity, we use \emph{figure extraction} to refer to the extraction of both figures and tables.
%in the remainder of the paper.

% One of the real-world applications of figure extraction from scanned ETDs would be to enhance the search engines for academic publications.
% This use case was demonstrated by Siegel et. al. in {\sc DeepFigures}~\cite{deepfigures} by deploying their system at scale on the Semantic Scholar website.

% Another potential application of this research is for mining data from the extracted figures from scanned ETDs.
% For example, an interesting use case would be finding the number of bar charts among the extracted figures, or perhaps even answering context-based questions such as `What is the peak value in a given bar plot in a figure?', and so on.

% Since many pre-existing works in the domain of document analysis cater only to born-digital documents~\cite{pdffigures2,deepfigures,tablebank,docbank}, our work which focuses on figures extraction from scanned documents brings us one step closer to tapping the vast knowledge-base of scanned documents.

There are many challenges to accurately identifying figures in scanned ETDs.
The image resolution and scanning quality may vary across the collection.
OCR output is often error-ridden.
Most older ETDs were typewritten. In very old documents, figures and tables may have been hand-drawn or rendered in a separate process and literally cut-and-pasted into typewritten documents.
Further, since ETD collections are cross-disciplinary, the documents in them present a variety of layout styles.

In the past decade, there has been a significant amount of effort on mining scholarly big data, represented by hundreds of millions of scholarly papers~\cite{khabsa2014plosone}.
Comprehensive frameworks were developed to segment scholarly papers into different levels of elements (e.g., GROBID~\cite{lopez2009grobid} and CERMINE~\cite{tkaczyk2015cermine}).
Table and figure extraction software also was developed~\cite{choudhury2013jcdl} (e.g., PDFFigures~\cite{clark15pdffigures}, PDFFigures2~\cite{pdffigures2}, TableBank~\cite{tablebank}, DocBank~\cite{docbank}, and {\sc DeepFigures}~\cite{deepfigures}).
However, these either rely on the underlying document structure of a PDF file~\cite{tkaczyk2015cermine,choudhury2013jcdl,clark15pdffigures,pdffigures2}, or exclusively cater to the analysis of born-digital documents~\cite{deepfigures,tablebank,docbank}.
%Further, some of them have been proposed for a specific of scholarly documents~\cite{choudhury2013jcdl}.

One of the models that inspired our work is {\sc DeepFigures}~\cite{deepfigures}, which generated high-quality labels for figures extracted in scientific documents.
{\sc DeepFigures} was trained on data derived from arXiv and PubMed datasets with a reported average precision of 96.8\%.
However, it performed poorly on scanned ETDs, as we demonstrate in the experiments below.

Following are some of the % major and novel 
contributions of our paper:
\begin{enumerate}
    \item We curate and release ScanBank, a new manually labeled dataset for figure extraction from scanned ETDs with 10K candidate page images labeled by humans as to the presence of the 3.3K figures or tables found therein~\cite{scanbank_dataset_zenodo}.
    % , which contains over 3.3K high quality figure labels spanning across more than 10K page images.
    \item We pose and answer important research questions aimed at finding better methods for figure extraction from scanned documents.
    \item We train and evaluate the performance of the YOLOv5 model to extract figures and tables from scanned ETDs. To the best of our knowledge, no existing baseline achieves comparable performance. 
    \item We propose novel data augmentation techniques to make born-digital documents look like scanned ETDs.
    \item We also release the source code used for producing the results in this paper, along with the trained models~\cite{source_code_zenodo}.
\end{enumerate}
% Our main contribution in this paper is demonstrating a deep neural model for accurately identifying figures and tables in scanned ETDs.
%Our technique takes the data used in the prior work to locate figures in born-digital scientific papers, but introduces various perturbations, or data augmentations, to the documents  to mimic scanned ETDs.

% --------------------------------------------------------------------
% --------------------------------------------------------------------
% --------------------------- Related work ---------------------------
% --------------------------------------------------------------------
% --------------------------------------------------------------------

\section{Related Work}
\label{sec:related_work}

In PDFFigures2~\cite{pdffigures2} proposed a new approach that analyzes the structure of individual pages by detecting chunks of body text, graphical elements, and captions -- and then locates figures and tables by reasoning about the empty regions in the pages.
Their results were used as a baseline in later work~\cite{deepfigures}.
PDFFigures2 was designed for born-digital PDF documents since it used a rule-based approach to extract figures by leveraging the underlying document structure of the PDF document, which is not explicit for scanned PDF documents.

\citeauthor{deepfigures} proposed {\sc DeepFigures}, a method for extracting figures, and tables from scholarly PDFs.
Data from arXiv\footnote{https://arxiv.org/help/bulk\_data} and PubMed\footnote{https://www.ncbi.nlm.nih.gov/pmc/tools/openftlist} were used for locating figures in scientific papers, which 
% For the arXiv dataset, the authors modified the original LaTeX source code of the PDFs such that it rendered visible bounding boxes around figures and captions.
% Figures were already identified in a parsed format of the PubMed dataset.
% Induced labels 
were then used for training a deep learning model, consisting of ResNet~\cite{resnet} in conjunction with the Overfeat architecture~\cite{overfeat}, to predict coordinates of the bounding boxes around figures.
% The figures were then extracted by cropping these predicted bounding boxes out from the rendered PDFs.
%Unlike PDFFigures2, DeepFigures does not use a rule-based approach, but its figure extraction performance decreases for scanned PDF documents (as shown in~Section~\ref{ss:exp_1}).

\cite{tablebank} proposed TableBank, an image-based table detection and recognition framework. % with a dataset.
Their contribution was a weakly supervised machine learning (ML) model trained on a dataset of Microsoft Word and LaTex documents crawled from the Web.
%TableBank contains documents 
The authors included documents of different languages (e.g., Chinese, English, Japanese, etc.) in TableBank, thereby making it more general.
% , which made their dataset more diverse.
The method also modified the document source code, allowing them to generate a ground truth dataset of figures and tables, with known bounding boxes.
The authors took a step further and built a recurrent neural network that converted a detected table (in an image format) into a table markup format (i.e., a table parsed into text).
In other words, it converted an image of a table into a structured machine readable format; this was called table structure recognition.

\cite{data_driven_recognition_and_extraction} employed an object detection model called Faster R-CNN, allowing them to achieve better region assignments for tables in a PDF document than {\sc DeepFigures}.
They introduced a dataset with 31,639 manually labeled PDF pages with image bounding boxes.
Like {\sc DeepFigures}, Faster R-CNN and TableBank were trained on born-digital documents. %, and hence their performance is expected to decrease for scanned documents.

Recently, \cite{library_of_congress_newspaper_navigator_paper} proposed the Newspaper Navigator dataset, used for extracting and analyzing visual content from 16 million historic newspaper pages from ``Chronicling America.''
This work used a manually labeled dataset of historic newspaper pages containing labels for seven classes (headlines, photographs, illustrations, maps, comics, editorial cartoons, and advertisements).
A pre-trained Faster R-CNN model was fine-tuned on this dataset to enable the extraction of the targeted visual content.
The documents used by this work were archived newspapers, which have different visual structures from scholarly documents, such as ETDs.

YOLO is a popular deep learning framework, designed to detect multiple objects in an input image in a single inference pass.
It is also well-known for its low space and time complexity during inference, which makes it an ideal alternative for deployment on devices where low resource consumption is vital.
The initial version of YOLO (YOLOv1) was proposed by \citeauthor{yolov1paper} in 2016.
This was the first work which, instead of repurposing classifiers as object detectors, framed object detection as a regression problem.
YOLOv1 detects multiple bounding boxes in a single forward pass, so it can be trained end-to-end directly for detection. 
Many subsequent versions of YOLO were proposed by various authors in the following years.
In 2020, the fifth version of YOLO (YOLOv5) was proposed \cite{yolov5github}, which achieved the best detection performance among the versions.
It has four different network sizes (small, medium, large, and extra-large), which allows users to make trade-offs between the time and space complexity.
%The exact details about the number of parameters and the run times for each size variant of YOLOv5 are documented~\cite{yolov5github}.
We adopt the extra-large version of YOLOv5 containing about 89 million trainable parameters.

Data augmentation~\cite{effectiveness_of_data_augmentation_in_classification} is a popular technique in deep learning that helps to train a model better without collecting new data.
% With this technique, the existing training data is modified and the deep learning model is re-trained on the modified data.
Some of the common data augmentation methods are affine transformations, random rotations, additive noise (e.g., salt-and-pepper, Gaussian), perspective transformations, and random cropping~\cite{data_augmentation_book}.
%A number of implementations have emerged online for implementing these techniques.
We use the popular ImgAug~\footnote{https://github.com/aleju/imgaug} open-source software, which provides the capability to augment not only the image but also the corresponding bounding boxes around images.
% For example, if an affine transformation is applied to the image, the same amount of affine transformation will be applied to the coordinates of the associated bounding boxes of that image. 

% In our experiments, we investigate whether the {\sc DeepFigures} performance could be improved by training on data augmented by various techniques. 
%To the best of our knowledge, such a study hasn't been done yet.

% --------------------------------------------------------------------
% --------------------------------------------------------------------
% --------------------------- Research questions ---------------------
% --------------------------------------------------------------------
% --------------------------------------------------------------------

\section{Research Questions}
\label{sec:research questions}
We address the following research questions (RQs) in this paper.

\textit{RQ1: How well can existing methods (e.g., {\sc DeepFigures}), extract figures from scanned ETDs?} Existing methods, like {\sc DeepFigures}, have been trained and tested exclusively on born-digital documents. Since scanned and born-digital documents differ considerably in visual appearance, it is necessary to investigate how well the existing methods for figure extraction perform on scanned ETDs.

\textit{RQ2: Can the performance of existing methods (e.g., {\sc DeepFigures}) be improved by training them using an augmented version of their original data or by weight initialization?} Since the visual appearance of scanned and born-digital ETDs differ significantly, we apply different data augmentation techniques on born-digital documents to make them visually look like scanned documents which are then used to train existing methods and evaluate their performance. Answers to RQ1 and RQ2 can help clarify if there is value in creating a dataset like ScanBank.
    
\textit{RQ3: Can figure-extraction performance for scanned ETDs be improved by training the DeepFigures or YOLOv5 model on our ScanBank dataset?} The original {\sc DeepFigures} model was trained using the born-digital arXiv dataset. The model in RQ2 was also trained using the born-digital arXiv dataset, albeit with augmentations. However, the source of the data in both cases is still born-digital and labels are automatically generated (Section~\ref{sss:deepfigures_label_induction}). Therefore, to answer this question, we train a state-of-the-art model (YOLOv5) using the ScanBank dataset.
%Therefore, whether the performance can be improved further by training on a dataset of scanned ETDs (labeled manually) still remains to be explored. This research question aims to address that. 
%\wu{I am not sure this is a a good question. What's the difference between manually labeled data with automatically generated data? These are born-digital, right? But we should be working on non-born-digital data, so why labeling born-digital data? }
%\sampanna{Hello Dr. Wu.  Yes, good point. I think a better way to write RQ3 could be as follows:  "Can this performance be improved by training on manually labeled scanned ETDs? The original DeepFigures model in~\cite{deepfigures} was trained using the arXiv dataset. Similarly, the model in RQ2 will also be trained using the arXiv dataset, albeit with augmentations. However, the data source for both of these models is still arXiv, which is a born-digital dataset and its labels are being automatically generated (see Section~\ref{sss:deepfigures_label_induction}). Furthermore, the visual appearance of the pages of a born-digital dataset, even after applying data augmentations, could be different than a dataset of scanned ETDs. Therefore, whether the performance can be improved further by training on a dataset of scanned ETDs (labeled manually) still remains to be explored. This research question aims to address that."}

\textit{RQ4: Can transfer learning improve performance, training only some of the layers of the deep neural network used in {\sc DeepFigures}?} We freeze some of the layers of the pre-trained deep neural network in {\sc DeepFigures} and re-train the model.

% --------------------------------------------------------------------
% --------------------------------------------------------------------
% --------------------------- Background -----------------------------
% --------------------------------------------------------------------
% --------------------------------------------------------------------

\section{Existing Frameworks}
\label{sec:background}

\subsection{DeepFigures}
\label{ss:approach_of_deepfigures}

Since our work builds upon {\sc DeepFigures}~\cite{deepfigures}, we review the general strategy employed in that work.
The overall approach of {\sc DeepFigures} is to generate high-quality labels for figure extraction in a large number of scientific documents for training a deep learning model (ResNet-101~\cite{resnet} + Overfeat~\cite{overfeat}).
% The generated dataset is then used to train a deep learning model which is a combination of ResNet-101~\cite{resnet} and the Overfeat~\cite{overfeat} architecture.

\subsubsection{Label induction}
\label{sss:deepfigures_label_induction}

This step of the {\sc DeepFigures} pipeline deals with two types of datasets.
The first is the arXiv\footnote{https://arxiv.org} dataset which can be obtained using AWS's S3 API\footnote{https://docs.aws.amazon.com/AmazonS3/latest/API/Welcome.html} as mentioned on their bulk access website\footnote{https://arxiv.org/help/bulk\_data}.
This dataset contains the LaTeX source code for each research paper in the dataset which is first used to compile the PDF which is converted to a list of page images.
Then the LaTeX source code is modified to add bounding boxes around figures.
This is achieved by adding a few lines of LaTeX code at the beginning of the source code.
These modified LaTeX files are compiled to generate PDFs that are converted to images.

Using this method, each scholarly paper in the dataset can be transformed into two lists of images.
The first does not have any bounding boxes around its figures but the second list does.
Each image in these two lists represents a single page from the PDF document.
Subtraction on each corresponding pair of images from these two lists in a pixel-by-pixel fashion yields a subset of the images, each of which has a bounding box around the figures.
The images resulting from the subtraction contain only the difference between the subtrahend and minuend, which are the bounding boxes around the figures.
The top-left and the bottom right of each of the bounding boxes is found using simple image processing operations and used as the co-ordinates of the bounding boxes.
% The connected components in the resultant images are used to find the coordinates of the bounding boxes.

Labels are also generated for the PubMed dataset using its auxiliary data which includes image files for all graphics and XML markup indicating the locations of captions.
These image files are used for multi-scale template matching on the page images of the original paper to obtain the figure locations which are then used as labels for training the {\sc DeepFigures} model.
The quality of a model is evaluated by manually labeling randomly sampled images and comparing against predicted labels.
% These bounding boxes are then used to train a deep learning object detection model (ResNet-101~\cite{resnet} + Overfeat~\cite{overfeat}).

\subsection{YOLOv5}
\label{ss:approach_of_yolov5}

Our proposed approach is based on YOLOv5~\cite{yolov5github}, which offers better usability with superior performance on the MS COCO benchmark~\cite{lin2014coco} compared with YoloV4.
The backbone network (for object detection) of YOLOv5 implements BottleneckCSP~\cite{wang2020cspnet}.
YOLOv5 chooses PANet~\cite{wang2019panet} for feature aggregation and adds an SPP~\cite{he2014spp} block after BottleneckCSP to increase the receptive field and separate out the most important features from the backbone~\cite{bochkovskiy2020yolov4}. 
In general, the models in the YOLO family have better performance and are more compact than models of similar or larger size.

\subsubsection{Data augmentation}
Before passing the input images to the model, YOLOv5 uses three methods to augment the data: scaling, color space adjustment, and mosaic augmentation.
The mosaic augmentation was a novel augmentation technique when YOLOv5 was developed, which works by combining four images into four tiles with random ratios~\cite{yolov5_article_roboflow}.

\subsubsection{Anchor boxes}
The YOLOv5 model predicts bounding boxes as deviations from a list of anchor boxes.
Using K-means and a genetic algorithm, an initial set of anchor bounding boxes is learned from the training set.
These anchor boxes are then used as references for learning the deviation to get the predicted bounding boxes~\cite{yolov5_article_roboflow}.

\subsubsection{Architecture}
% As mentioned above, the backbone of the YOLOv5 model is based on 
Cross Stage Partial (CSP) networks~\cite{wang2020cspnet} used in YOLOv5 have significantly lower number of trainable parameters and use fewer flops since they address the problem of duplicate gradients from larger convolutional networks~\cite{yolov5_article_roboflow} resulting in a faster inference time for YOLOv5.
% , when performance is crucial (Figure~\ref{fig:yolov5_architecture}).

% \begin{figure}[ht]
%     \centering
%     \includegraphics[width=\linewidth]{figures/YOLOv5 model architecture without dimensions.png}
%     \caption{Architecture of YOLOv5 (extra-large)~\cite{yolov5github}}
%     \label{fig:yolov5_architecture}
% \end{figure}

\subsection{Comparison of YOLOv5 and {DeepFigures}}
\label{ss:comparison_of_yolov5_and_deepfigures}

\subsubsection{Parameters vs. inference times}
\label{sss:parameter_vs_inference_times_DF_YOLOv5}

Table~\ref{tab:comparison_of_deepfigures_and_yolov5} compares the {\sc DeepFigures} and YOLOv5 models on the number of trainable parameters, the average inference time across 500 images processed on an Nvidia Tesla P100 GPU, and the year the model was proposed.
The inference times in Table~\ref{tab:comparison_of_deepfigures_and_yolov5} do not include any time needed for pre-processing the image or post-processing the predictions.
We measure only the time to make a single forward pass for a single pre-processed image on the GPU.

\begin{table}[htb]
	\centering
	\caption{Comparison between {\sc DeepFigures} and YOLOv5.}
	\begin{tabular}{cccc}
		\toprule
		Model           & \# params  & Inference time & Year published  \\
		\midrule
		DeepFigures     & 45 million & 33.514 ms & 2018 \\
		YOLOv5 (XL)     & 89 million & 35.048 ms & 2020 \\
		\bottomrule
	\end{tabular}
	\label{tab:comparison_of_deepfigures_and_yolov5}
\end{table}

The inference times are almost the same even though the number of parameters in YOLOv5 is almost double that of {\sc DeepFigures}, which is because of the CSP network.

\subsubsection{Architectural comparison}
\label{sss:architectural_comparison_DF_YOLOv5}

The backbone of the {\sc DeepFigures} model is ResNet-101, and the head uses the Overfeat architecture.
The ResNet backbone consists of several CNN layers stacked on top of each other with skip connections between them.
With these skip connections, an entire copy of the previous layer's outputs is bypassed and added to the current layer's outputs.
This makes the back-propagation of gradients much easier for deep layers, and significantly reduces learning parameters and convergence time for deep neural networks.

The backbone and the head of YOLOv5 consist of a number of stacked 1D and 2D convolution, up-sampling, and BottleNeckCSP layers.
BottleNeckCSP layers are BottleNeck layers that incorporate the CSP architecture.
In this architecture, a part of the outputs of the base layers is split into two parts.
The first part is directly linked to the end of the stage, thereby skipping the model stage.
The second part goes through the model stage and is then concatenated together with the first part.
% The CSP architecture can be applied to any staged neural network such as ResNet, DenseNet, and BottleNeck~\cite{wang2020cspnet}.

Apart from these architectural differences, YOLOv5 also performs mosaic data augmentation, which is absent in {\sc DeepFigures}.
Furthermore, YOLOv5 has adaptive bounding box anchors which are absent in {\sc DeepFigures}.
These anchors help with better prediction in case the classes to be detected have a bias for shapes.
%For example, if we are detecting two classes, i.e. pens and soccer balls, most of the bounding boxes for pens will be elongated while the bounding boxes of soccer balls will be more square.

% --------------------------------------------------------------------
% --------------------------------------------------------------------
% --------------------------- Data -----------------------------------
% --------------------------------------------------------------------
% --------------------------------------------------------------------

\section{Data}
% In this work, we use two datasets: the arXiv dataset and the MIT dataset. 
\subsection{arXiv dataset}
The arXiv dataset consists of the LaTeX source code of research papers.
The authors of {\sc DeepFigures}~\cite{deepfigures} used these LaTeX files to induce labels for figures and then trained their models based on these labels.
We augment the data to make it visually similar to scanned data, used for training models.
% Then we use this augmented data to train the model.
Our proposed augmentation techniques are elaborated in Section~\ref{sec:data_augmentation}.

The arXiv dataset is born-digital by nature, which is still different from scanned (non-born-digital) ETDs.
This is why the MIT dataset is introduced.

\subsection{MIT dataset}
As opposed to the arXiv dataset, PDFs in the MIT dataset are not compiled from LaTeX source but by scanning physical hard-copies of ETDs.
However, because of a lack of bounding box information, we need to manually label PDFs in this dataset.
The result is ScanBank, a manually labeled subset of the MIT dataset, described in Section~\ref{gold_standard_creation}.
We use it for training and evaluation.

\subsection{Characteristics of scanned PDFs}

%\wu{I feel this paragraph is redundant.}
%In~\cite{deepfigures}, the labeled data is generated from two datasets, i.e. arXiv and PubMed. In the case of arXiv, the LaTeX source of each paper is used to generate the labels. This means that the compiled PDF is `born-digital' and not scanned (for example using a flat-bed scanner). Similarly, for the PubMed dataset, most of the PDFs used to generate the labels are born-digital. However, for the older PDFs, although they are scanned, the figure/caption markup in the auxiliary data is not reliable, sometimes non-existent. For example in~\cite{pubmed_old_paper_example_1,pubmed_old_paper_example_2,pubmed_old_paper_example_3,pubmed_old_paper_example_4}), although we can see the figures in the PDF, they are not always present in the markup. This implies the presence of noise in the PubMed training data.

% This means that the distribution of the training data for the DeepFigures model primarily consists of born-digital PDFs. Naturally, DeepFigures therefore tries to mimic the same distribution as it's training data.

As opposed to born-digital PDFs, scanned PDFs originally existed as hard-copies and were later digitized into PDF using scanning tools such as flat-bed scanners.
The process of scanning introduces certain artifacts in PDFs.
For example, the content for some pages might be slightly rotated or tilted because of errors in the placement of paper in the scanner.
Other kinds of noise such as salt-and-pepper noise, blurring, and perspective transformations are also possible.
The content of the PDFs could have been typed using a typewriter, or even hand-written.
Therefore, the overall appearance of a scanned PDF can vary significantly from a born-digital PDF.
As a result, the feature distribution of the data on which {\sc DeepFigures} was trained is significantly different from that of scanned PDFs.
This leads to worse performance for {\sc DeepFigures} for scanned ETDs, according to our experiments. 
%Quantifiable results to support this argument are presented later in this paper.

% Point out that DeepFigures focused on any scholarly document (especially research papers), but our experimentation is done mostly on ETDs. Also point out why we focus only on ETDs, and why any model which works on research papers can also work on ETDs.

% In the DeepFigures paper, they had to add white borders to the original latex source code because the coloured borders were moving the figures by a few pixels. In our work, we are trying to change the font size of the documents to 

% --------------------------------------------------------------------
% --------------------------------------------------------------------
% --------------------------- Methods -----------------------------------
% --------------------------------------------------------------------
% --------------------------------------------------------------------

\section{Proposed Methods}
\label{sec:methods}

\subsection{The ScanBank dataset}
\label{gold_standard_creation}
To overcome the limitation of this born-digital dataset, we create a new non-born-digital benchmark standard dataset for evaluation.
% To act as a reference dataset for evaluating the models, we create our novel ScanBank dataset to act as a reference dataset for evaluating our models.
% The data generated using our data augmentation pipeline proposed in Section~\ref{sec:data_augmentation} is one of the datasets used to train our models.
% However, we create our novel ScanBank dataset to act as a reference dataset for evaluating our models.

\subsubsection{Collecting ETDs}
We download the PDFs and metadata of all ETDs from MIT's DSpace repository\footnote{https://hdl.handle.net/1721.1/7582}.
% \cite{mit_dspace_main_page}.
We choose this source for three reasons.
First, all of these ETDs were initially submitted as paper copies and scanned into PDF.
Second, the ETDs are organized by department\footnote{https://dspace.mit.edu/} which facilitates sampling over different fields of study.
Third, each ETD has associated metadata that can be used for sampling across years.

\begin{figure}
    \centering
    \includegraphics[width=\linewidth]{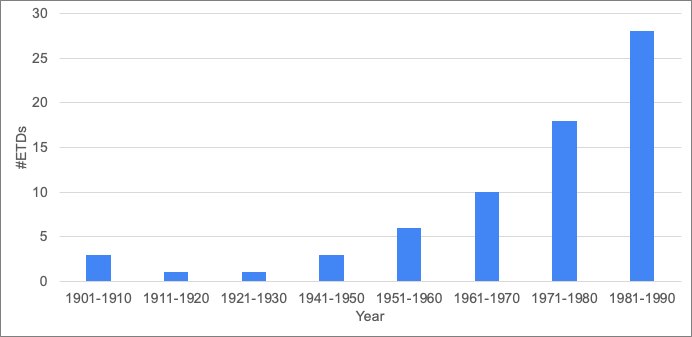}
    \caption{Year-wise distribution of ETDs sampled for the ScanBank dataset.}
    \label{fig:gold_standard_yearwise_stats}
\end{figure}

From the downloaded ETDs, we randomly sampled ETDs with the following constraints.
The publication date needs to be prior to 1990.
At most one ETD comes from each sub-community (doctoral, master's, and bachelor's) within each department.
After accounting for empty sub-communities, our sample contained a total of 70 ETDs (see Figure~\ref{fig:gold_standard_yearwise_stats}).

\subsubsection{Labeling ETDs}
%Next, we converted each page from PDF to an image, and scaled the resolution to 100 dots per inch.
%For example, if a PDF document of an ETD contained 100 pages, then after converting each page of the PDF document into an individual image, we will be left with a total of 100 images.
%Each image resembles a ``screenshot'' of a page from the PDF.
Next, we converted each page from PDF to an image.
Thus, if an ETD had 100 pages, we would have a total of 100 images.
Then, we scaled the resolution to 100 dots per inch, so an image resembles a ``screenshot'' of a page from the PDF.
We did not change the original aspect ratio of the pages when converting them to images.

A total of 10,182 images of pages were obtained across the 70 sampled ETDs.
We used the VGG Image Annotator (VIA)~\cite{via_paper,via_website} to manually label these images with bounding boxes around figures.
The VIA tool provides a graphical user interface for manual labeling of images.
We used rectangular bounding boxes whose coordinates can be recorded using mouse click-and-drag in the VIA tool.
Each bounding box contains the coordinates of the top left corner of the bounding box and its actual width and height in pixels.

The following labeling guidelines were used: 
%\wu{I don't quite understand why table of content and list of figures are labeled.}
%\sampanna{In this work, we are trying to extract figures (i.e. figures and table). Since the table of contents and the list of figures visually look like a structured table, I decided to label them.}

\begin{enumerate}
    \item Some ETDs contained source code snippets. These code snippets were not labeled.
    \item ``Table of Contents,'' ``List of Figures,'' and ``List of Tables'' sections were labeled since they are visually similar to tables.
    \item Captions for both figures and tables were labeled.
    \item The bibliography was not labeled since it can neither be classified as a figure nor a table.
    \item Math equations (including matrices) were not labeled.
    \item For screen-captures (including newspaper clippings that contain figures), the individual figures within the figures were labeled. The encompassing figure was not labeled. No nested or overlapping labelling was done.
\end{enumerate}

When manually labeling the ScanBank dataset, the captions for respective figures were included in their respective bounding boxes to be consistent with the choice of {\sc DeepFigures}. In total, 3,375 figures were labeled across the entire dataset.
%In addition, it is more convenient for future tasks to extract captions from figures/tables from an already extracted sub-region of the page than from the entire page.

%\wu{Why are there 10100 images? from 3375 figures? }
%\sampanna{Good catch. That was a mistake. I have removed that phrase. Probably happened when copy-pasting text.}
%\sampanna{Hello Dr. Wu. Are you referring to the training process which uses this dataset? It is described in the experimental setup subsections of the individual experiments. Did I understand your question correctly?}
    
% \begin{figure}
%     \centering
%     \includegraphics[width=0.5\textwidth]{figures/gold_standard_figures_per_etd_2.png}
%     \caption{Distribution of figure count per ETD in the gold standard dataset.}
%     \label{fig:gold_standard_figures_per_etd}
% \end{figure}

\subsubsection{Validation and test splits\label{ss:gold_standard_val_test_split}}
We split the ScanBank dataset into two equal halves after shuffling it randomly.
The first half will be used as the validation set for fine-tuning or choosing the best model during training.
The second half will be used as the test set for evaluating the model.

\subsubsection{Accessing ScanBank}
The ScanBank dataset is freely available online~\cite{scanbank_dataset_zenodo}.
This contains a \textit{.json} file which contains the coordinates of the 3.3K bounding boxes that represent the figures in the 10K images.
Further, to limit the size of the downloaded file, this dataset only contains the URLs of the ETDs which were used to create the 10K images in this dataset.
The Python source code and instructions, which are included in the dataset, can be executed to download the ETDs and convert them into the individual 10K page images.

\subsection{Data augmentation\label{sec:data_augmentation}}

\begin{figure*}
\begin{minipage}{0.33\textwidth}\centering
\includegraphics[width=\textwidth]{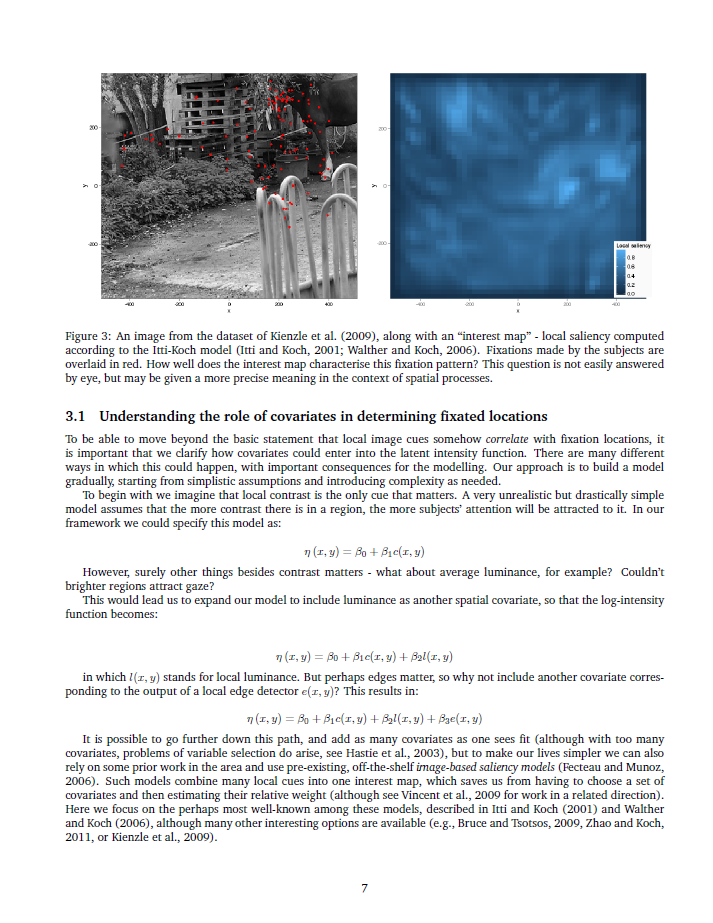}
\end{minipage}
\begin{minipage}{0.33\textwidth}\centering
\includegraphics[width=\textwidth]{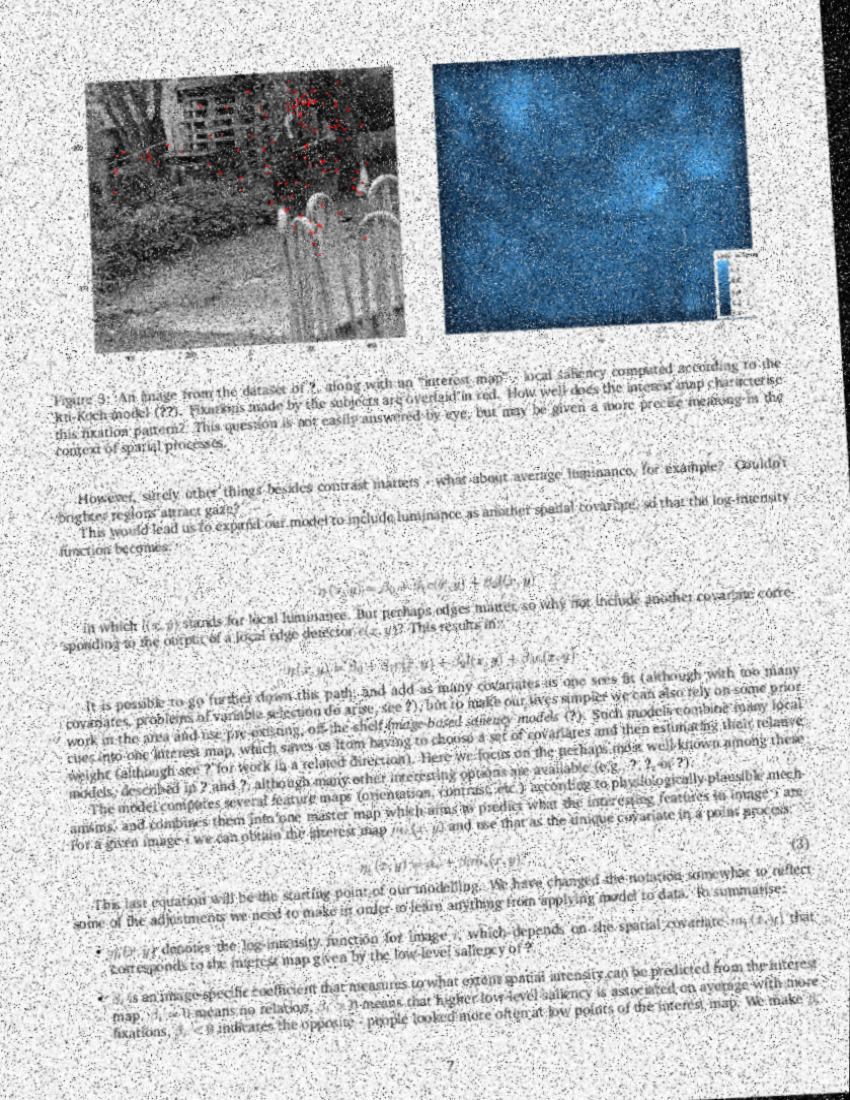}
\end{minipage}
\begin{minipage}{0.33\textwidth}\centering
\includegraphics[width=\textwidth]{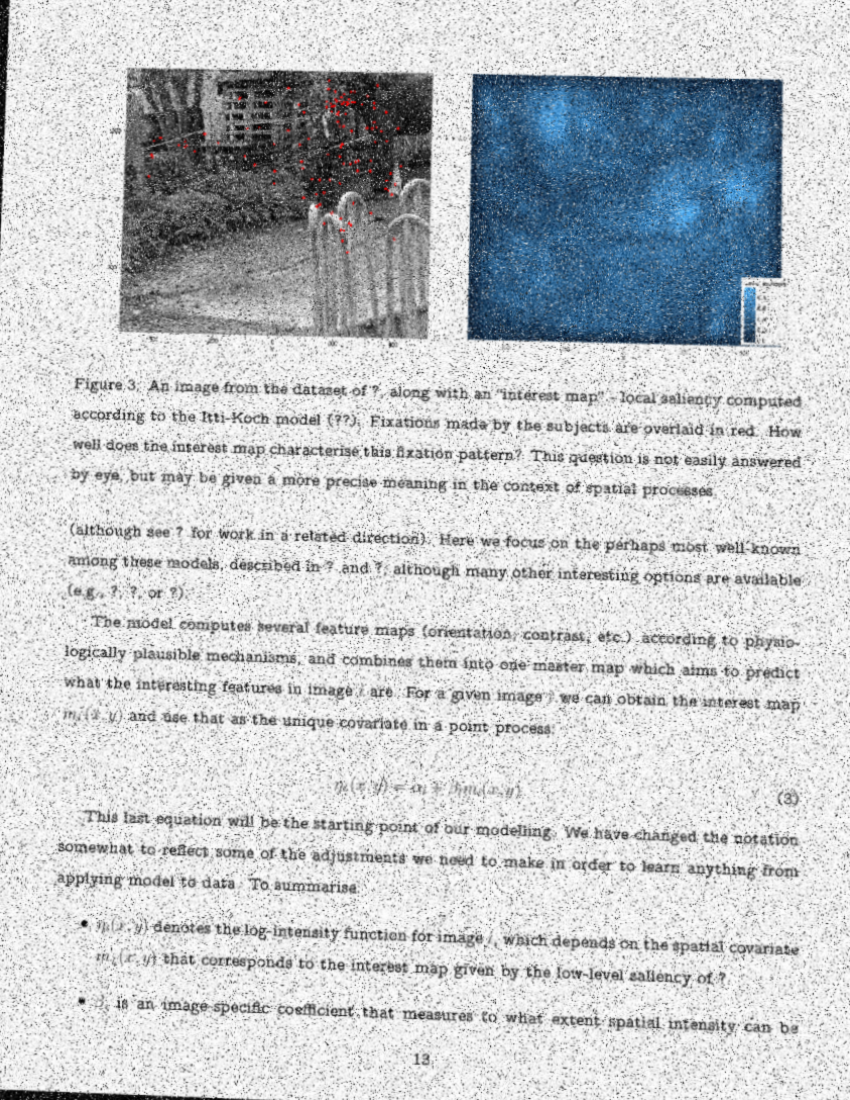}
\end{minipage}
\caption{
    (a) The left image shows the original page from~\cite{imgaug_sample_pdf} (page number 7). 
    (b) The middle image shows the same page after applying only image-based transformations. 
    (c) The right image shows the page after applying both image-based and LaTeX-based transformations. Note the change in font-size and font layout.
}
\label{fig:aug_results}
%\Description[The three sub-figures show the effects of latex-based augmentations]
%{
%    (a) The left image shows the original page from~\cite{imgaug_sample_pdf} (page number 7).
%    (b) The middle image shows the same page after applying image-based transformations. This image does not contain the latex-based transformations.
%    (c) The right image shows the page after applying latex-based transformations. This image contains both image-based and latex-based transformations. Note the change in font-size and font layout.
%}
\end{figure*}

%We propose to improve the results of {\sc DeepFigures} by augmenting the training data using various data augmentation strategies.
We propose to improve the performance of {\sc DeepFigures} by applying data augmentation strategies to the training data.
The purpose is to apply data augmentation techniques on born-digital documents with the purpose of making them resemble scanned documents..
% This can be achieved by modifying the original training data such that its visual appearance resembles that of scanned documents.
We use image-based and LaTeX-based transformations to modify the original documents. 
\subsubsection{Image-based transformations}\label{sss:image_based_transformations}
We use a software library called ImgAug\footnote{https://github.com/aleju/imgaug} to apply image-based transformations on each page.
Each transformation below is available as a function in ImgAug.

\paragraph{Random affine rotation} 
While scanning the hard-copy of a document, pages may be slightly rotated, and hence might not be perfectly aligned..
% As a result, in the scanned PDF file, not all pages will be perfectly aligned, and could be tilted by random angles.
Therefore, we rotate each page of a PDF file by \textit{n} degrees, where \textit{n} is a float sampled from a standard uniform distribution.
% randomly sampling a number 
% Therefore, for each page of a PDF file, we randomly sample a floating-point number between $-5$ and $+5$ using standard normal distribution.
% We rotate the image of this page by that number of degrees.

\paragraph{Additive Gaussian noise} A flatbed scanner works by reflecting the light from paper and creating an image of the paper based on the naturally reflected light.
%Light reflecting from the surface of the paper is a natural phenomenon.
Hence, we use Additive Gaussian Noise to mimic this effect.
The parameters of this noise are heuristically chosen using trial-and-error.

\paragraph{Salt-and-pepper noise}
Salt-and-pepper noise is often seen on images caused by sharp and sudden disturbances in the image signal~\cite{salt_and_pepper_noise_book}.
We heuristically chose 0.1 as the probability of replacing a pixel by noise.

\paragraph{Gaussian blur}
% Different from 
Unlike natural (analog) images, digital images must be encoded with a specified resolution resulting in a pre-determined number of bytes, and some loss of sharpness.
%Digital images are limited to a certain resolution as compared with natural objects.
%This happens because to represent any image in its digital form, there are only a certain number of bytes available to encode it.
%Therefore, during the process of digitizing any hard-copy, a certain sharpness/resolution is lost.
Therefore, we apply Gaussian blurring to smoothen the images using a Gaussian Kernel ($\sigma=0.5$).
% \begin{equation}
%     G(x,y) = \frac{1}{2\pi\sigma^2}e^\frac{-x^2-y^2}{2\sigma^2} \label{eq:gaussian_kernel}
% \end{equation}
% with size set to $\sigma=0.5$.

\paragraph{Linear contrast} 
Although today's scanners are built using modern technology, they are incapable of capturing all colors of a natural object.
To incorporate this scanning effect, we add Linear Contrast (parameterized by $\alpha=1$).
% , parameterized by $\alpha$, which determines the amount of linear contrast to introduce in the image. We use $\alpha=1$.

\paragraph{Perspective transform}
Since scanned pages can sometimes look stretched, we implement Perspective Transform\footnote{https://imgaug.readthedocs.io/en/latest/source/overview/geometric.html} in which a part of the image (formed by randomly selecting 4 points) is stretched.
% Scanned pages sometimes look stretched.
% Accordingly, we implement Perspective Transform in which four points are randomly selected from the four corners of the image and the distance of these four points from the respective corners of the image is sampled randomly using a normal distribution.
% The random distribution can be controlled using the scale factor, where the scale roughly measures how far the perspective transformation's corner points may be distanced from the image's corners\footnote{https://imgaug.readthedocs.io/en/latest/source/overview/geometric.html}.
% ~\cite{imgaug_perspective_transform}.

Random affine rotation and perspective transform might cause geometric changes in the images which could lead to a mismatch between the locations of bounding boxes and locations of the figures.
To correct this mismatch, we use a feature in ImgAug that transforms the bounding boxes according to the transformations being applied.

Figure~\ref{fig:aug_results} (middle) shows the image of a page from a research paper from arXiv~\cite{imgaug_sample_pdf} after applying the image-based data augmentation operations mentioned above.
% The original page can be seen in Figure~\ref{fig:aug_results} (left).

\subsubsection{LaTeX-based data augmentations}\label{sss:latex_based_transformations}
Although image-based transformations change the overall appearance of a page, they do not change the inherent structure of the text within the page.
The text still is well-formed and does not quite resemble the text from a PDF scanned from a typewritten document.
To achieve such effects, we impose modifications in the LaTeX source code in addition to the image-based transformations.

\begin{figure}[ht]
    \centering
    \includegraphics[width=0.45\textwidth]{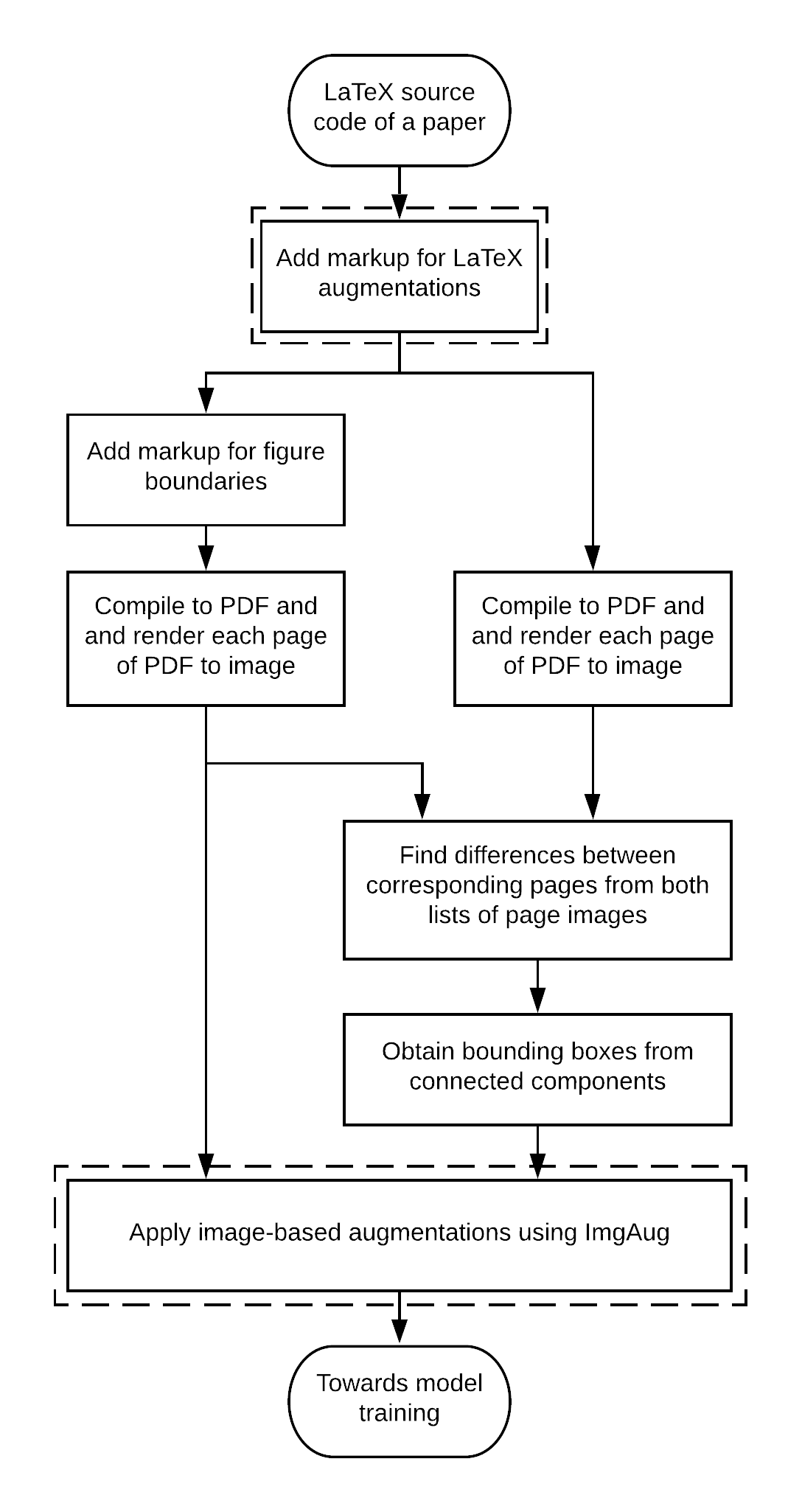}
    \caption{Our proposed pipeline to obtain training labels using data augmentation. Our proposed modifications on top of {\sc DeepFigures} ~\cite{deepfigures} are highlighted using dotted lines.}
    %Modified on top of {\sc DeepFigures} ~\cite{deepfigures}. Our proposed modifications are highlighted using dotted lines.}
    \label{fig:deepfigures_modified_pipeline_flowchart}
\end{figure}

\paragraph{Change font size} We incorporate this effect by modifying the following command at the beginning of each document.
    \begin{verbatim}
    \documentclass[sigconf]{acmart}
    \end{verbatim}
    We replace it with the following command:
    \begin{verbatim}
    \documentclass[sigconf,12pt]{acmart}
    \end{verbatim}
    
\paragraph{Modify LaTex macros} We add the following LaTeX source code before the beginning of the document to change the font type of the document to look more like a typewriter font and also increases the line spacing to 1.5:
    \begin{verbatim}
    \renewcommand\ttdefault{cmvtt}
    \renewcommand{\familydefault}{\ttdefault}
    \linespread{1.5}
    \end{verbatim}
% This changes the font type of the document to look more like a typewriter font and also increases the line spacing to 1.5.

Figure~\ref{fig:aug_results} (right) shows the image after applying LaTeX-based transformations.
The change in the font type and text layout as compared to the original page is shown in Figure~\ref{fig:aug_results} (left).
% It should be noted that any changes in the page layout arising because of the LaTeX-based augmentations do not decrease the label quality.
% This is because the layout changes occur in both of the images being subtracted.
% Therefore, the only differences between the subtrahend and minuend are the bounding boxes around the figures.
The overall processes for obtaining the labels from these augmented images are illustrated in Figure~\ref{fig:deepfigures_modified_pipeline_flowchart}; they occur after incorporating the proposed modifications.

% --------------------------------------------------------------------
% --------------------------------------------------------------------
% ------------------------------- Results------- ---------------------
% --------------------------------------------------------------------
% --------------------------------------------------------------------

\section{Results}
\label{sec:results}

\subsection{Experiment 1: Evaluating the pre-trained {\sc DeepFigures} model using ScanBank}
\label{ss:exp_1}

\subsubsection{Experimental setup}
\label{sss:exp_1_setup}

\citeauthor{deepfigures} released their source code and model weights used for training {\sc DeepFigures}.
%However, {\sc DeepFigures}  was originally trained on born-digital scholarly documents.
In this experiment, we evaluate its performance on our proposed ScanBank dataset consisting of scanned ETDs.
We run inference for {\sc DeepFigures} (using the trained model weights released in~\cite{deepfigures}) on the validation and test splits of ScanBank.

\begin{table}[ht]
	\centering
	\caption{Performance of {\sc DeepFigures}  on 4 Datasets. }
	\label{tab:deepfigures_performance_on_gold_standard}
	\begin{tabular}{l*{6}{c}r}
	\toprule
		{\bf Model}    & {\bf Precision} & {\bf Recall}  & {\bf F1} \\
		\midrule
		CS-Large & - & - & 0.849 \\
		PubMed &  - & - & 0.806 \\
		\midrule
		Validation & 0.461 & 0.491 & 0.475  \\
		Testing  &  0.439 & 0.445 & 0.442  \\
	\bottomrule
	\end{tabular}
\end{table}

\subsubsection{Evaluation metrics}
\label{sss:exp_1_metrics}

The {\sc DeepFigures} model outputs a set of bounding boxes for each figure it detects in the page image.
We filter out the bounding boxes whose confidence scores are lower than 0.5.
We match the predicted bounding boxes with the true bounding boxes to minimize the total Euclidean distance between the centers of paired bounding boxes.
This is an instance of the linear assignment problem~\cite{deepfigures,pdffigures2} solved using the Hungarian algorithm~\cite{hungarian_algorithm_paper}.
% We match the predicted bounding boxes with the true bounding boxes to minimize the total Euclidean distance between the centers of paired bounding boxes, which is an instance of the linear assignment problem described in detail in~\cite{deepfigures}.
Once the predicted boxes have been matched with the true boxes, we deem a predicted box as correct if its intersection over union (IOU) with the true box is greater than or equal to 0.8 (true positive), incorrect if less than 0.8 (false positive)~\cite{deepfigures}.
When a ground truth is present in the image and the model fails to detect it, we deem it a false negative.
Using these  metrics, we calculate the precision, recall, and F1 scores.
The choices of the thresholds described in Section~\ref{sss:exp_1_metrics} are made to be consistent with choices made in {\sc DeepFigures}~\cite{deepfigures}.

% \subsubsection{Justification of using these evaluation metrics}
% \label{sss:exp_1_metrics_justification}
% The choice of the various thresholds described in Section~\ref{sss:exp_1_metrics} is dependent on similar choices made in {\sc DeepFigures}~\cite{deepfigures}.
% In {\sc DeepFigures}, an IOU threshold of 0.8, a confidence threshold of 0.5, precision, recall, and F-1 scores were used to describe their model's performance.
% Therefore, to maintain parity and for a fairer and easy comparison, we too choose the same metrics and thresholds for our experiments.

\subsubsection{Results}
\label{sss:exp_1_results}
Table~\ref{tab:deepfigures_performance_on_gold_standard} shows the performance of {\sc DeepFigures} evaluated on four datasets.
CS-Large and PubMed datasets are born-digital scholarly documents. (Only the F1 scores were reported.)
The ``Validation'' and ``Testing'' are scanned ETDs.
The performance of {\sc DeepFigures} with scanned ETDs is considerably lower than its performance on CS-Large and PubMed datasets, indicating that it is not suitable for accurately extracting their figures and tables.

\subsection{Experiment 2: Ablation studies}
\label{ss:exp_2}

\subsubsection{Motivation}
\label{sss:exp_2_motivation}

In Section~\ref{sec:data_augmentation}, we described the different proposed data augmentation transformations, which were chosen heuristically.
% Thus, it is important t
To know the effectiveness of these transformations
% To this end,
we conduct ablation studies
in two parts --- first for LaTeX-based transformations and then for image-based transformations.
% In each of these, we do a leave-one-out ablation study.

In our implementation, each transformation in the list mentioned in Section~\ref{sss:image_based_transformations} and Section~\ref{sss:latex_based_transformations} can easily be disabled or enabled.
The total number of possible combinations for this is $2^n$, where \textit{n} is the number of transformations possible.
Training $2^n$ models for a significantly large number of transformations may not always be computationally feasible.
Therefore, we propose a leave-one-out ablation study to get a better idea about which transformations are actually helpful.

\subsubsection{Leave-one-out ablation study}
\label{sss:leave_one_out_albation_study}

% In a leave-one-out ablation study, w
We train $n$ different models in parallel ($n$ is the total number of possible transformations).
% In each of these models
For each model, we keep all of the hyper-parameters constant, except the list of transformations to apply.
For the $i$-th model, we disable the $i$-th transformation and leave the remaining enabled.
Based on the results, if the $i$-th model performs poorly, we can say that disabling the $i$-th transformation has worsened the performance and therefore, enabling that transformation contributes positively towards the model performance.

Although the leave-one-out ablation study gives us some idea about the performance of each transformation, it does not give the entire picture.
For instance, two transformations performing well on their individual ablation studies could perform worse when used together.
This can be further extended for combinations of more than two transformations.
Therefore, we do not claim that the leave-one-out strategy gives us the best combination of transformations.
However, it is helpful for weeding out the transformations that led to a poor performance.

\subsubsection{Results}
\label{sss:exp_2_results}

We trained 11 deep learning models described in Section~\ref{sss:leave_one_out_albation_study} for 24 hours. 
% The results are displayed in Table~\ref{tab:ablation_study_result}.
For each model, we choose the checkpoint that performed the best on the validation set of ScanBank.
Then, we reported its performance on the test set (Table~\ref{tab:ablation_study_result}).
The performance of the original pre-trained {\sc DeepFigures} model was computed by directly running inference on our testing dataset.
%If a model selected using this method performs well on the test set, we can safely say that it will perform well on any new data it encounters. \wu{I am not sure if this is true. This largely depends on whether the new data have the same distribution with the test data, right? }
%\sampanna{You are right Dr. Wu. Maybe I need to rephrase this line. What do you think about the following? "If a model selected using this method performs well on the test set, we can safely say that it will perform well on any new data it encounters, given that it follows a distribution similar to the val and test sets."}

\begin{table}[htb]
	\centering
	\caption{Performance of the original {\sc DeepFigures}  model on the ScanBank dataset compared to the models trained as part of our leave-one-out ablation study. Inside parentheses are transformations disabled and the training time in hours.}
	\begin{tabular}{ccccccc}
		\toprule
		{\bf Model} & {\bf Precision} & {\bf Recall} & {\bf F1} \\
		\midrule
		DeepFigures                        & 0.450 & 0.468 & 0.459 \\
		\midrule
		Ours (All enabled, 24)             & 0.550 & 0.521 & 0.535 \\
		Ours (All enabled, 72)             & 0.556 & 0.498 & 0.526 \\
		\midrule 
		Ours (Gaussian Noise, 24)          & 0.547 & 0.456 & 0.498 \\
		Ours (Gaussian Noise, 72)          & 0.578 & 0.492 & 0.531 \\
		\midrule
		Ours (Affine, 24)                  & 0.521 & 0.496 & 0.508 \\
		Ours (Affine, 72)                  & 0.591 & 0.457 & 0.516 \\
		\midrule
		Ours (Gaussian Blur, 24)           & 0.450 & 0.471 & 0.460 \\
		Ours (Gaussian Blur, 72)           & 0.557 & 0.542 & 0.550 \\
		\midrule
		Ours (Linear Contrast, 24)         & 0.559 & 0.527 & 0.542 \\
		Ours (Linear Contrast, 72)         & 0.567 & 0.488 & 0.524 \\
		\midrule
		Ours (Perspective Transform, 24)   & 0.526 & 0.517 & 0.521 \\
		Ours (Perspective Transform, 72)   & 0.431 & 0.586 & 0.497 \\
		\midrule
		Ours (Salt and Pepper, 24)         & 0.574 & 0.579 & 0.576 \\
		Ours (Salt and Pepper, 72)         & 0.554 & 0.525 & 0.539 \\
		\midrule
		Ours (Line spacing 1.5, 24)        & 0.454 & 0.642 & 0.532 \\
		Ours (Line spacing 1.5, 72)        & 0.542 & 0.641 & 0.588 \\
		\midrule
		Ours (Typewriter font, 24)         & 0.543 & 0.465 & 0.501 \\
		Ours (Typewriter font, 72)         & 0.624 & 0.490 & 0.549 \\
		\bottomrule
	\end{tabular}
	\label{tab:ablation_study_result}
\end{table}

We make the following observations from Table~\ref{tab:ablation_study_result}. (1) The F1-scores of almost all of our models surpass the F1-score of the original {\sc DeepFigures}  model, indicating that the performance of figure extraction for scanned documents can be improved  using augmented data. (2) The model in which Gaussian Blur was disabled has a score of 0.460.
This is close to the F1-score of the original {\sc DeepFigures}  model (0.459).
In all of the other models, Gaussian Blur was enabled, and their performance is significantly higher than the original {\sc DeepFigures}  model.
Apparently, Gaussian Blur is the most helpful transform.

To find out how the model performance changes with training time, we re-run this experiment by re-training the eleven models for 72 hours.
% , the results of which are also tabulated in Table~\ref{tab:ablation_study_result}.
Again, we observe that almost all of our models significantly out-perform the original {\sc DeepFigures}  model, which is consistent with the observations when models were trained for 24 hours (Table~\ref{tab:ablation_study_result}).
However, the new experiments show contrastive results when the models are trained for 72 hours.
For example, the F1-score of the model in which Gaussian Blur is disabled is no longer the model with the lowest performance.
In contrast, it significantly outperforms the model when the training time is 24 hours.
On the other hand, the F1-score of the model in which Additive Gaussian Noise is disabled (0.531) is higher than the F1-score of the model in which Affine transform is disabled (0.516).
The F1 decreases the most when Perspective Transform was disabled, indicating that Perspective Transform seemed the most helpful. 
Because of these contrasting observations, the current experiments do not prove which type of transformation is the most helpful.
The significant change of F1 when the training time is 72 hours indicated the model at 24 hours may not be sufficiently trained.
Training for a longer time is needed to observe the decrease of F1 for certain features before conclusively assessing their importance in data augmentation.

\subsection{Experiment 3: Training YOLOv5}
\label{ss:exp_3}

\subsubsection{Experimental setup}
\label{sss:exp_3_setup}

Experiments~1 and 2 used the {\sc DeepFigures}  model.
However, the best F1-score we have obtained is still less than 0.588.
In this experiment, we use YOLOv5, a state-of-the-art object detection framework.
% As mentioned in Section~\ref{sec:related_work}, w
We choose the extra-large version of YOLOv5 with about 89 million trainable parameters (Section~\ref{sec:related_work}).
We train on the ScanBank dataset with batch size of eight, and use eight-fold cross-validation to report its performance.
%We use a batch size of eight.

\subsubsection{Results}
\label{sss:exp_3_results}

% The results of the eight-fold cross-validation are in Table~\ref{tab:yolov5_cross_val_result}.
We observe that the mean F1-score of all the cross validation folds is 0.860 with a standard deviation of 0.073 (Table~\ref{tab:yolov5_cross_val_result}).
This F1-score is significantly higher than the F1-score of the original {\sc DeepFigures} model evaluated on our ScanBank standard dataset (Table~\ref{tab:deepfigures_performance_on_gold_standard}).
This F1-score is also significantly higher than the F1-score obtained in any of our previous experiments.

In the previous experiments, we used the {\sc DeepFigures} model architecture which uses a combination of ResNet-101 and Overfeat.
The total number of parameters in {\sc DeepFigures} is about 45 million.
YOLOv5's model contains about 89 million parameters.
This means that the number of trainable parameters in YOLOv5 is almost double that of {\sc DeepFigures} (Table~\ref{tab:comparison_of_deepfigures_and_yolov5}).
This result is consistent with previous observation that empirically, a model's performance increases with the model size \cite{huang2019gpipe}. 
%\wu{This needs to be justified more carefully. What about overfitting? } \sampanna{Isn't cross-validation a robust counter-measure against overfitting? Apologies if I understood your question incorrectly.}

\begin{table}[htb]
	\centering
	\caption{Performance of the original {\sc DeepFigures}  model on the ScanBank dataset compared with the 8-fold cross validation of YOLOv5.}
	\begin{tabular}{cccc}
		\toprule
		 Model                         & Precision & Recall  & F1 \\
		\midrule
		DeepFigures                    & 0.450 & 0.468 & 0.459 \\
		\midrule
		YOLOv5 (K=0)                   & 0.749 & 0.869 & 0.804 \\
        YOLOv5 (K=1)                   & 0.870 & 0.821 & 0.845 \\
        YOLOv5 (K=2)                    & 0.75 & 0.691 & 0.720 \\
        YOLOv5 (K=3)                   & 0.928 & 0.972 & 0.949 \\
        YOLOv5 (K=4)                   & 0.886 & 0.937 & 0.911 \\
        YOLOv5 (K=5)                    & 0.887 & 0.935 & 0.910 \\
        YOLOv5 (K=6)                   & 0.804 & 0.889 & 0.844 \\
        YOLOv5 (K=7)                   & 0.859 & 0.932 & 0.894 \\
		\midrule
		YOLOv5 (Mean)                   & 0.842 & 0.881 & 0.860 \\
		YOLOv5 (Std. dev.)              & 0.066 & 0.090 & 0.073 \\
		\bottomrule
	\end{tabular}
	\label{tab:yolov5_cross_val_result}
\end{table}

\subsection{Experiment 4: Comparison with other models}
\label{ss:exp_4}

\subsubsection{Experimental setup}
\label{sss:exp_4_setup}

In this section, we compare our results with other models on this problem.
In our work, we try to extract figures from scanned ETDs.
Other similar models were not exactly designed for this task.
Therefore, we compare our results with other research in this field which aims to achieve similar goals.
For example, Google Cloud's commercial machine learning offering AutoML, and Microsoft Azure's Custom Vision, allow users to upload a labelled dataset and train it on the cloud without the need to select any model architecture or hyper-parameters.

We uploaded our labeled ScanBank dataset to Google Cloud and Microsoft Azure and trained models on these platforms. 
For Google Cloud AutoML, we let the platform choose how to split the dataset into different classes for training. Such an option is not available for Microsoft Custom Vision, so we used 80\% of the dataset for training and validation, and the rest for testing.
Amazon AWS's SageMaker (AutoPilot) supports only Regression, Binary Classification, and Multiclass Classification\footnote{https://docs.aws.amazon.com/sagemaker/latest/dg/autopilot-automate-model-development-problem-types.html}.
% Another promising cloud platform is Amazon AWS, which offers SageMaker with AutoPilot.
% However, AutoPilot supported only three problem types: Regression, Binary Classification, and Multiclass Classification\footnote{https://docs.aws.amazon.com/sagemaker/latest/dg/autopilot-automate-model-development-problem-types.html}. %\cite{aws_sagemaker_autopilot}.
Since it does not yet support Object Detection, we exclude it from this experiment.
Another baseline we compare is proposed by~\cite{library_of_congress_newspaper_navigator_paper}.
In their work, a Detectron2 model is trained to extract figures  from a manually labelled dataset of historic scanned American newspapers, the visual appearance of which significantly differs from scanned ETDs.
%However, it is somewhat similar to ours because it too uses a scanned dataset. Further,~\cite{library_of_congress_newspaper_navigator_paper} performs figure extraction on this dataset, which is also something that we have proposed in our work.
We used the pre-trained model released in~\cite{library_of_congress_newspaper_navigator_paper} to run inference on our ScanBank dataset.
%Since this model was already pre-trained, we used the entire gold standard dataset for running inference for performance evaluation.
%We then compared these predictions with the labels from the gold standard dataset to evaluate the performance.
%The pre-trained model predicts bounding boxes for scanned newspaper pages.
Each predicted bounding box is labelled with one of the following seven classes: Photograph, Illustration, Map, Comics/Cartoon, Editorial Cartoon, Headline, and Advertisement.
The last two classes (i.e., Headline, and Advertisement) are not figure-like.
Therefore, for the purpose of this experiment, we only consider the predictions for the first five classes, and apply non-maximal suppression to eliminate duplicate predictions (confidence/objectness threshold = 0.5, IOU threshold = 0.8).

For all of these models, we used a confidence threshold of 0.5 to filter out less confident predictions.
Further, to compare the predicted labels with ScanBank's labels, we used an IOU threshold of 0.8 to maintain parity with our previous experiments and with {\sc DeepFigures}.

% Cloud-based models are generally considered state-of-the-art.

\subsubsection{Results}
\label{sss:exp_4_results}

% Table~\ref{tab:comparison_results} shows the performance of the various models.
The performance of the Newspaper Navigator model \cite{library_of_congress_newspaper_navigator_paper} is significantly lower than the YOLOv5 model trained on our ScanBank dataset (Table~\ref{tab:comparison_results}).
This is likely because the Newspaper Navigator model was originally trained on a different dataset, while AutoML and Custom Vision models were trained on the ScanBank dataset.
% Meaning, although the Newspaper Navigator model was trained to extract figures using a scanned dataset (which is what we are doing in Section~\ref{ss:exp_3}), there was still a significant performance drop.
Moreover, our ScanBank dataset includes the labels for tables too, which the Newspaper Navigator model was not explicitly trained to extract.
This drop in performance further highlights the novelty and distinct use-case satisfied by our ScanBank dataset.
The drop in performance could also be potentially explained by the different model architectures used (i.e., YOLOv5 vs. Detectron2).
However, such a significant drop in performance for similar tasks is highly unlikely in two state-of-the-art models from a similar time-period.

Since Custom Vision does not disclose the architecture of its model(s), it is difficult to investigate its performance trends.
However, it was surprising that even when trained on our ScanBank dataset, 
% Regardless of that, unlike the pre-trained Newspaper Navigator model (which has never been trained on the gold standard dataset), Azure's Custom Vision model was indeed trained on the gold standard dataset.
% Further, cloud-based ML offerings from industry-leading cloud providers are generally considered state-of-the-art.
% % Further, it is also a common notion that commercial ML cloud products from industry-leading cloud providers would often outperform contemporary models trained by individuals.
% However, it can be observed from Table~\ref{tab:comparison_results} that 
Custom Vision performs significantly lower than the YOLOv5 model trained in Section~\ref{ss:exp_3}.
Similar to Custom Vision, AutoML does not disclose the specifics of its models.
However, it is mentioned that AutoML\footnote{https://www.fast.ai/2018/07/23/auto-ml-3/} uses Neural Architecture Search (NAS) to automatically find the best model architecture for the given task, which could be one of the reasons for higher performance of AutoML on ScanBank (Table~\ref{tab:comparison_results}).
However, NAS usually is more computationally expensive than a regular fixed-architecture neural network.
% Further, the research and technology used in AutoML could be more up-to-date than the one used in YOLOv5 (which is a few months older than AutoML) which could be another reason for AutoML's higher performance.
% Lastly, AutoML could have used an ensemble of morels for training and making predictions thereby giving it an edge over YOLOv5, but also making it further computationally expensive.

\begin{table}[htb]
	\centering
	\caption{Performance comparison of various models with 8-fold cross validation of YOLOv5 trained in Section~\ref{ss:exp_3}.}
	\begin{tabular}{cccc}
		\toprule
		Model                           & Precision & Recall & F1 \\
		\midrule
		DeepFigures                     & 0.450 & 0.468 & 0.459 \\
		\midrule
		Newspaper Navigator (LOC)       & 0.328 & 0.311 & 0.320 \\
		Azure Custom Vision             & 0.468 & 0.564 & 0.511 \\
        Google AutoML                   & 0.908 & 0.878 & 0.893 \\
		\midrule
		YOLOv5 (trained on ScanBank)                   & 0.842 & 0.881 & 0.860 \\
		\bottomrule
	\end{tabular}
	\label{tab:comparison_results}
\end{table}

% TODOs:
% 1. Reword the title of the table, and the names of the models it needed. In general, beautify the table a bit, if needed.
% 2. How much of the above should be done in the discussion section?

% --------------------------------------------------------------------
% --------------------------------------------------------------------
% -------------------------- Discussion ------------------------------
% --------------------------------------------------------------------
% --------------------------------------------------------------------

\section{Discussion}
\label{sec:discussion}

In Section~\ref{ss:exp_1}, we evaluated the performance of the pre-trained {\sc DeepFigures}  model on ScanBank (Table~\ref{tab:deepfigures_performance_on_gold_standard}) which served as the baseline for subsequent experiments.
% In the results of this experiment, w
We observed that the F1-score of the pre-trained {\sc DeepFigures}  model was substantially lower for extracting figures from scanned scholarly documents than that of born-digital ones.
This is likely due to the different visual characteristics of a scanned document and a born-digital document 
% . There differences are due to the artifacts 
introduced during scanning hard copies.
% , e.g., the way the document was written (typewriter, handwritten), paper color, etc.

\textbf{Answer to RQ1}: The performance of the original {\sc DeepFigures}  is significantly lower for figure extraction from scanned ETDs as compared to its performance for figure extraction from born-digital documents.
% For figure extraction from scanned ETDs, it achieves an F1-score of 0.459.

In Section~\ref{ss:exp_2}, we use the various data augmentation techniques described in Section~\ref{sss:image_based_transformations} and Section~\ref{sss:latex_based_transformations} to improve the performance of {\sc DeepFigures}.
The goal of these data augmentation techniques is to leverage the LaTeX source code of the training data to make the compiled PDFs look more like scanned PDFs.
% From the results i
In Table~\ref{tab:ablation_study_result}, we observe that models trained using augmented data almost always produced a higher F1-score than the original pre-trained {\sc DeepFigures}  model, indicating the effectiveness of our data augmentation techniques.
% were effective in bringing the visual appearance of the born-digital documents closer to that of scanned documents.

\textbf{Answer to RQ2}: The original {\sc DeepFigures}  model can be improved by retraining it on augmented data using weight initialization from the pre-trained model.

In Section~\ref{ss:exp_3}, we train YOLOv5 to improve the F1-score further.
% In this experiment, w
We initialized the weights randomly since we did not have any pre-trained set of YOLOv5 weights for a similar task.
When we trained the YOLOv5 (extra-large) on the ScanBank dataset using 8-fold cross-validation, we obtained a mean F1-score of 0.86, indicating the advantages introduced by YOLOv5 and its relatively big size compared to {\sc DeepFigures} (see Table~\ref{tab:comparison_of_deepfigures_and_yolov5}).
A similar trend was seen in improvement of image classification tasks with larger models \cite{huang2019gpipe}.
%The huge difference of the number of trainable parameters between YOLOv5 and DeepFigures can magnify the difference in the F1-score of these models, especially when the size of the training set is not relatively small.

% It is possible that the loss function of {\sc DeepFigures} got stuck in a local minimum because of the weight initialization.
% The optimizer used makes it difficult for the model to escape out of a local minimum thus preventing it from achieving a global minimum on the augmented data.

The better performance of YOLOv5 compared with {\sc DeepFigures}  could also be attributed to the mosaic data augmentation in YOLOv5.
Although our data augmentation techniques try to make the pages look more like scanned pages, the mosaic data augmentation has been shown to provide a stronger regularization effect \cite{bochkovskiy2020yolov4}.
Furthermore, the backbones used in these two networks are different which contribute to the difference in their performance.
YOLOv5's backbone heavily borrows from CSPNet while {\sc DeepFigures}  uses the ResNet-101.
% The different backbones could also contribute to the difference in their performance.

\textbf{Answer to RQ3}: The performance of the original {\sc DeepFigures}  model was not improved by training on manually labeled data. However, by using YOLOv5, we were able to achieve an F1-score much higher than any of the trained {\sc DeepFigures}  models.

All models in Section~\ref{ss:exp_2} were trained on the augmented born-digital arXiv dataset.
To check whether the performance of {\sc DeepFigures}  can be further improved, we conducted two more experiments.
In both of these experiments, we initialized the weights of the {\sc DeepFigures}  model with the weights released in the original {\sc DeepFigures}  paper~\cite{deepfigures} and trained the models on the ScanBank dataset with an 80-20 split.
In the first model, we allowed all layers to train, while in the second model, we allowed only the final fully-connected layers to train (a.k.a., the Overfeat layers).
% During the training, we periodically ran inference on the test set to track the performance of the models.
In both experiments, the F1-score decreased and never surpassed the original score.

\textbf{Answer to RQ4}: The performance of the original {\sc DeepFigures}  model was not improved by using transfer learning.

% --------------------------------------------------------------------
% --------------------------------------------------------------------
% -------------------------- Conclusion ------------------------------
% --------------------------------------------------------------------
% --------------------------------------------------------------------

\section{Conclusion and Future Work}
\label{sec:conclusion}

This work focuses on extracting figures from scanned ETDs.
We introduce our ScanBank dataset, which, to the best of our knowledge, is the first manually annotated dataset for figure and table extraction from scanned ETDs.
In our ablation study, some of our augmentation methods did not help (e.g. Salt-and-pepper and line spacing 1.5), others resulted in F1-scores even higher than the pre-trained {\sc DeepFigures} model (e.g. Gaussian Blur).
Table~\ref{tab:ablation_study_result} shows the results of our leave-one-out ablation study. 
% In our ablation study, disabling most of our proposed data augmentation methods still resulted in F1-scores higher than that of the pre-trained {\sc DeepFigures}  model.
Finally, the YOLOv5 model trained on the ScanBank dataset beats all of the previous models by a significant margin.
%We hope that our work will help identify and extract figures from the vast amount of scanned ETDs.

One of the real-world applications of this type of research would be to enhance the search engines for academic publications.
This was demonstrated in {\sc DeepFigures} by deploying their system at scale on the Semantic Scholar website.
Another potential application is for visual question answering from the extracted figures.
Examples include finding the number of bar charts from the extracted figures, or answering questions from the extracted figures, such as ``What is the peak value in a given plot?''
Our work can be potentially used for building search interfaces of archived periodicals in digital libraries such as HathiTrust~\cite{downie2020hathitrust}. 
%Since many relevant works in this domain cater to data extraction from born-digital documents~\cite{pdffigures2,deepfigures,tablebank,docbank}, our work which focuses on figures extraction brings us one step closer to tapping the vast knowledge-base of scanned documents.

%\section{Future Work}
%\label{sec:future_work}
% In the future, w
We plan to boost the performance by doubling the size of the ground truth dataset and combining heuristic methods (as in PDFFigures2) and learning based methods.
An alternative and promising approach to generate scanned ETDs out of born-digital ETDs is to leverage the recent advances in style transfer using CycleGANs~\cite{cycle_gan_style_transfer_paper}.
We will also investigate differential performance of the proposed model on figures and tables, respectively, given separate labels for these two content types.
Figures and tables extracted from our model can be used for building figure search functionalities for large scale digital library search engines for ETDs.

\bibliographystyle{unsrtnat}
\bibliography{references}  %%% Uncomment this line and comment out the ``thebibliography'' section below to use the external .bib file (using bibtex) .





% \end{thebibliography}

\end{document}